\documentclass[twoside]{article}

\usepackage{aistats2018}

\usepackage[utf8]{inputenc}
\usepackage{wrapfig}
\usepackage{subcaption}
\usepackage{amsmath}
\usepackage[colorinlistoftodos]{todonotes}

\DeclareMathOperator*{\argmin}{arg\,min}

\begin{document}
\twocolumn[
\aistatstitle{Data-Free Knowledge Distillation\\for Deep Neural Networks}
\aistatsauthor{Raphael Gontijo Lopes \And Stefano Fenu \And Thad Starner}
\aistatsaddress{Georgia Institute of Technology \And Georgia Institute of Technology \And Georgia Institute of Technology}]

\begin{abstract}
Recent advances in model compression have provided procedures for compressing large neural networks to a fraction of their original size while retaining most if not all of their accuracy. However, all of these approaches rely on access to the original training set, which might not always be possible if the network to be compressed was trained on a very large dataset, or on a dataset whose release poses privacy or safety concerns as may be the case for biometrics tasks. We present a method for data-free knowledge distillation, which is able to compress deep neural networks trained on large-scale datasets to a fraction of their size leveraging only some extra metadata to be provided with a pretrained model release. We also explore different kinds of metadata that can be used with our method, and discuss tradeoffs involved in using each of them.
\end{abstract}

\section{INTRODUCTION}
It is widely understood that training larger deep neural networks almost uniformly yields better accuracy on a variety of classification tasks than networks with fewer parameters. This has led to it becoming increasingly commonplace to train networks with incredibly large numbers of parameters, which can make deployment or model-sharing impractical or costly.

However, once a model is trained, a big architecture may not be required to represent the function it has learned \cite{hornik1991approximation}. Even without radical architectural changes, experimental approaches to model compression have shown 40x decreases in the memory profile of existing networks without significant loss of accuracy \cite{han2015deep_compression,han2015learning}. Therefore it is clear that the problem of model compression is of great practical interest, as it allows one to train huge models in data-centers, then compress them for deployment in embedded devices, where computation and memory are limited.

\begin{figure}[h]
\begin{center}
\includegraphics[width=0.3\textwidth]{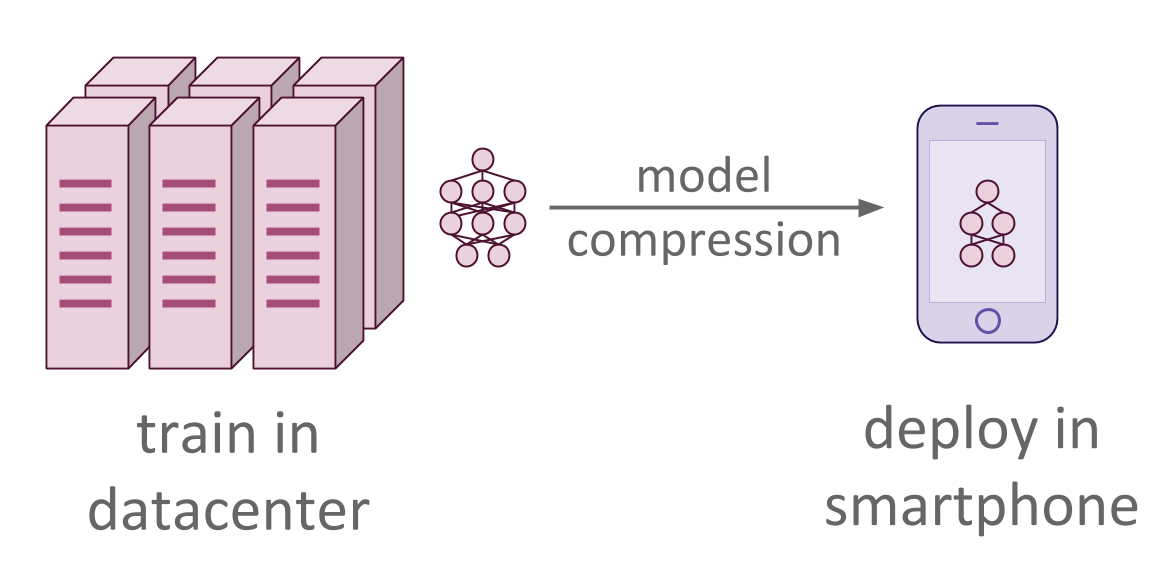}
\end{center}
\caption{A production pipeline for Deep Learning models: an over-parameterized model is trained to high accuracies using the computation power in a data center, then is compressed for deployment in a smartphone.}
\end{figure}

One issue with existing approaches is that they frequently require access to the original training data. As datasets get larger, their release can become prohibitively expensive. Even when a big dataset is released \cite{abu2016youtube}, it usually represents only a small subset of a much larger internal dataset used to train many state of the art models. Additionally, many datasets encounter hurdles to release in the form of privacy or security concerns, as in the case of biometric and sensitive data. This can complicate both data-sharing by the original trainer and data-collection by the model compressor.

This paper aims to explore the following question: ``Can we leverage metadata about a network to allow us to compress it effectively without access to its original training data?". We present a novel neural network compression strategy based on knowledge distillation \cite{hinton2015distilling} that leverages summaries of the activations of a network on its training set to compress that network without access to the original data.

\section{RELATED WORK}

Most compression methods for neural networks fall into three major camps: weight quantization, network pruning, and knowledge distillation. All of these methods work fairly independently of each other, and can be combined in different ways to get 35x-49x reductions in the memory profile of state of the art models \cite{han2015deep_compression}.

Weight quantization attempts to compress a network at the level of individual neuron weights, keeping all the parameters of a network and simply attempting to represent each individual parameter using less space. State of the art weight quantization methods can lead to very high classification accuracy even just with two \cite{rastegari2016xnor} or three \cite{li2016ternary} bits per parameter.

Network pruning, originally presented in Lecun's Optimal Brain Damage paper \cite{lecun1989optimal} attempts to directly reduce the number of parameters by zeroing out individual neuron weights entirely. Further work showed that not only is it a valid way to decrease the overall memory profile of a network, but that it is also a good way to deal with overfitting \cite{hassibi1993optimal} and can help a network better generalize. Recent methods showed that there are pruning methods that can compress a network without any loss of accuracy \cite{han2015learning}.

\begin{figure}[h]
\begin{center}
\includegraphics[width=0.4\textwidth]{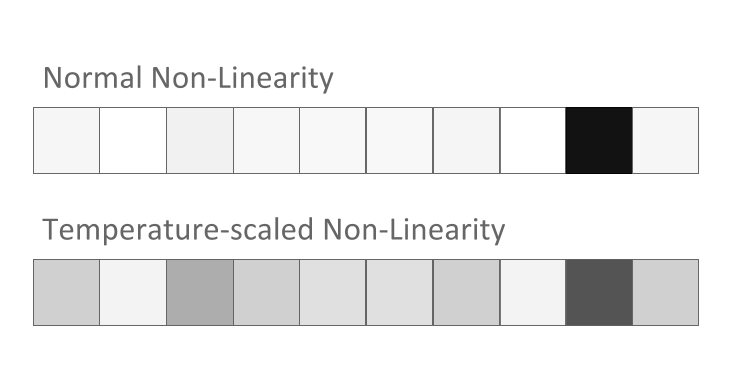}\end{center}
\caption{The effect of scaling non-linearities to some temperature: the softened activations should provide the student model with more information about how the teacher model generalizes.}
\end{figure}

Instead of attempting to prune the weights or neurons of a network, another branch of approaches, termed ``Knowledge Distillation" train a smaller ``student" network to copy the actions of a larger ``teacher" network. This is typically done either by attempting to train a shallow student network \cite{ba2014deep, bucilua2006model} or a thin one \cite{romero2014fitnets} to match the outputs of the ``teacher" network. A generalized approach, from which we draw heavily, was proposed by Hinton et al. \cite{hinton2015distilling}: it relies on modifying the last layer of the teacher network so that, instead of outputting a classification, it outputs activations scaled to some temperature parameters in an attempt to provide more information about how the teacher model generalizes.

The modifications to Hinton's knowledge distillation method which we propose are in part inspired by Intrinsic Replay, introduced by \cite{draelos2016neurogenesis} as a way to re-stabilize information a network had already learned after its architecture changed to compensate for new information. In our compression method, instead of relying on the original dataset to guide the optimization of the student network, we attempt to regenerate batches of data based on metadata collected at training time describing the activation of the network. We also depart from \cite{draelos2016neurogenesis} in that Intrinsic Replay requires a generative model to recreate the data used to retrain the network, where our method can be applied to any convolutional or fully-connected classifier by simply regenerating samples of the input using the inversion method proposed in \cite{mahendran2015understanding}.

Since this method works independently of any sort of quantization or pruning, it is possible to use it as part of a preliminary step to the sort of compression pipeline proposed in \cite{han2015deep_compression}. We also posit that it could be used as an alternative to the retraining step after each pruning iteration in scenarios where there's no access to the original training set.

\begin{figure}[h]
\begin{center}
\includegraphics[width=0.49\textwidth]{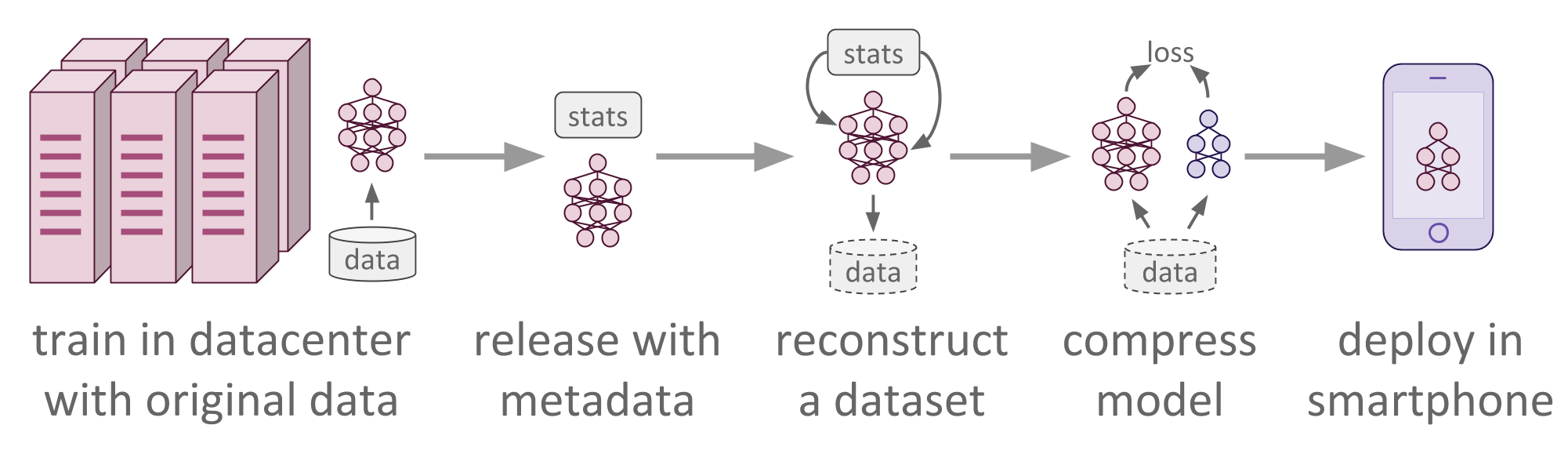}
\end{center}
\caption{The proposed model compression pipeline: a model is trained in a datacenter and released along with some metadata. Then, another entity uses that metadata to reconstruct a dataset, which is then used to compress the model with Knowledge Distillation. Finally, the model is deployed in a smartphone.}
\end{figure}

\section{METHOD}

After training the teacher network on the original dataset, we compute records for the activations of each layer in the network, and save those alongside the model. These can take different forms, and several such forms are discussed in more detail in Section \ref{stats_section}.

In order to train the student without access to the original data, we attempt to reconstruct the original dataset using only the teacher model and its metadata in the form of precomputed activation records. This is done similarly to \cite{mahendran2015understanding}, attempting to find the set of images whose representation best matches the one given by the network. We pass random gaussian noise as input to the teacher model, then apply gradients to that input noise to minimize the difference between the activation records and those for the noise image. Doing this repeatedly allows us to partially reconstruct the teacher model's view of its original training set.

More formally, given a neural network representation $\phi$, and an initial network activation or proxy thereof $\phi_0=\phi(x_0)$, we want to find the image $x^*$ of width $W$ and height $H$ that:
$$x^* = \argmin_{x \in R^{HxW}} \; l(\phi(x), \phi_0) $$
where $l$ is some loss function that compares the image representation $\phi(x)$ to the target one $\phi_0$. Unlike the work in \cite{mahendran2015understanding}, we use loss functions that attempt to use precomputed activation statistics for $\phi$ to find images that maximize the similarity between the neural response of the network for the original dataset and that for the reconstructed dataset instead of trying to maximize some classification probability. These are detailed more clearly in \ref{stats_section}. Once we have a reconstructed dataset, it can be fed as training input to the student network without any further modification.

\begin{figure*}[!ht]
\begin{center}
\begin{subfigure}{0.48\textwidth}
\centering
\includegraphics[width=\textwidth]{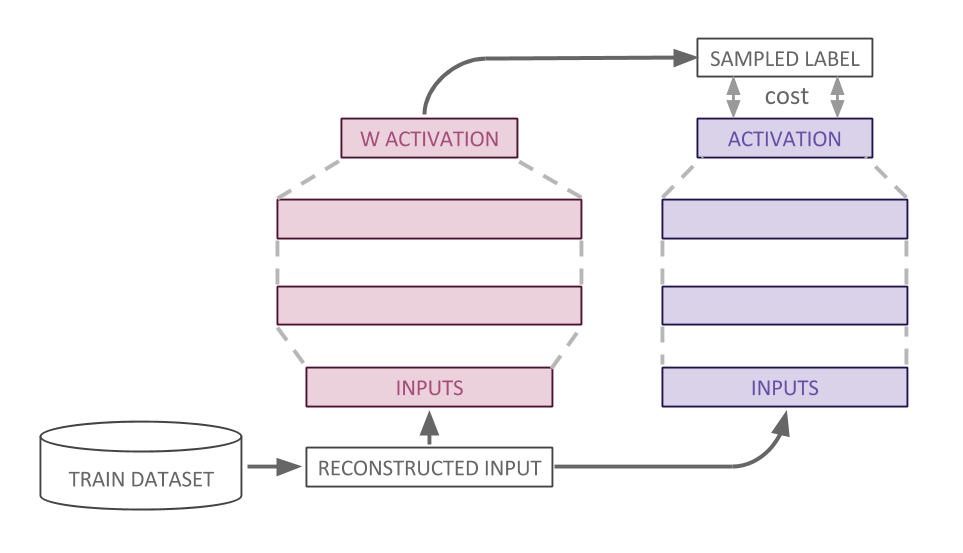}
\subcaption{Knowledge distillation}
\label{pure_distill}
\end{subfigure}

\begin{subfigure}{0.43\textwidth}
\centering
\includegraphics[width=\textwidth]{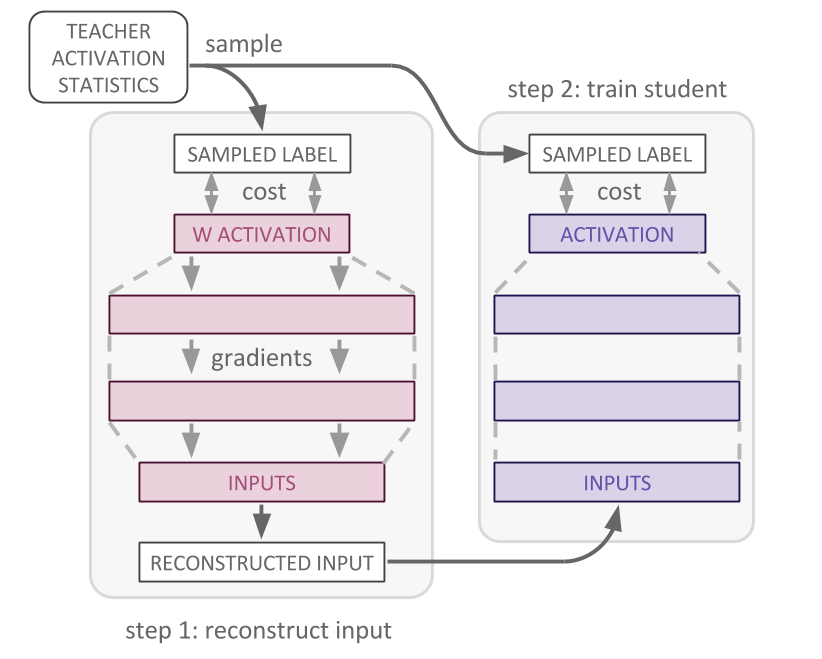}
\subcaption{Top Layer Activation Statistics}
\label{top_layer_diagram}
\end{subfigure}
\begin{subfigure}{0.56\textwidth}
\centering
\includegraphics[width=\textwidth]
{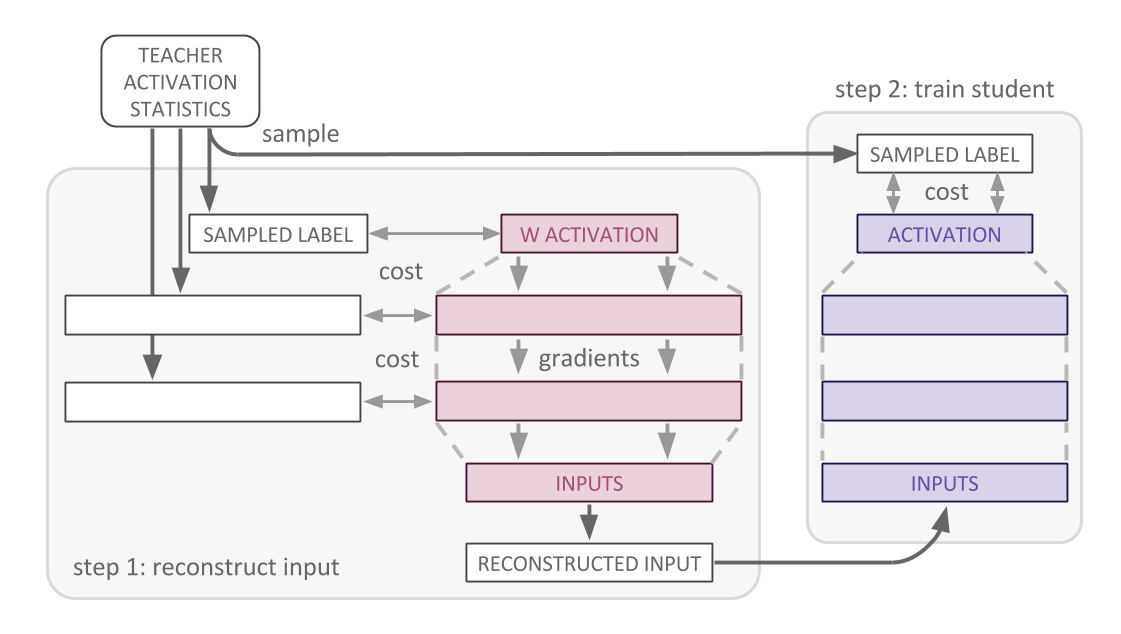}
\subcaption{All layers Activation Statistics}
\label{all_layers_diagram}
\end{subfigure}

\begin{subfigure}{0.496\textwidth}
\centering
\includegraphics[width=\textwidth]
{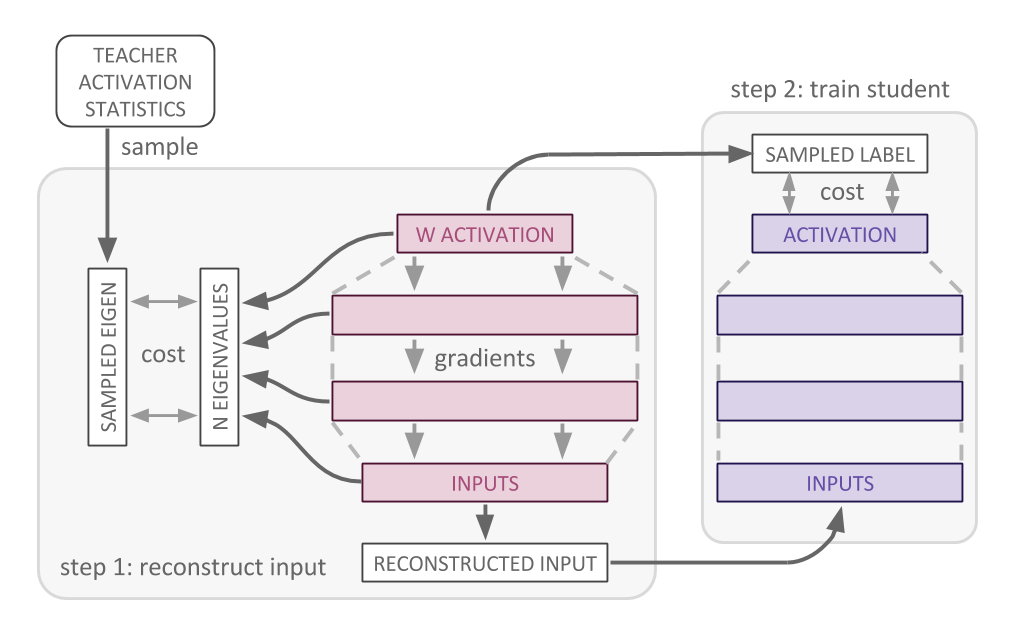}
\subcaption{All-Layers Spectral Activation Record}
\label{spectral_all_layers}
\end{subfigure}
\begin{subfigure}{0.496\textwidth}
\centering
\includegraphics[width=\textwidth]
{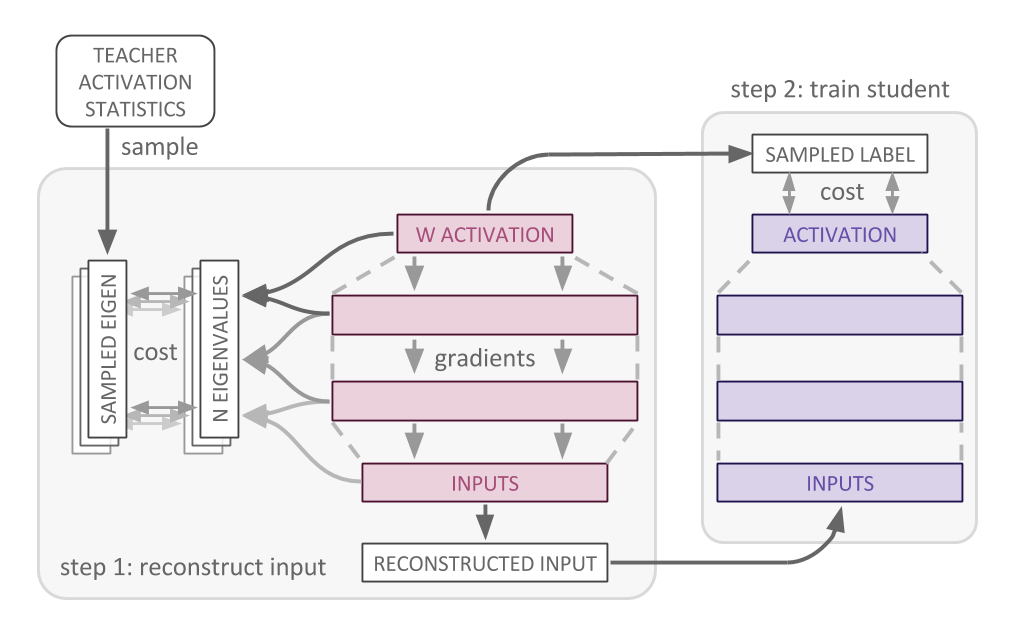}
\subcaption{Layer-Pairs Spectral Activation Record}
\label{spectral_pair_layers}
\end{subfigure}

\caption{Overview of the different activation records and methods used to reconstruct the original dataset and train the student network. In (\subref{pure_distill}), the student network is trained directly on examples from the original dataset as input, and the teacher's temperature-scaled activations as labels. In (\subref{top_layer_diagram}), we keep activation statistics for the top layer of the teacher network. Then, we sample from those, and optimize the input to the teacher to recreate those activations. That reconstructed input is then used to train the student network. (\subref{all_layers_diagram}) is very similar to (\subref{top_layer_diagram}), but it involves recording statistics, sampling, and recreating activations for all layers of the network. In (\subref{spectral_all_layers}), the optimization objective is to reconstruct the entire activation of the network to correspond to a compressed version of the original network activation. This is intended to better capture inter-layer dynamics and is initially done by expanding the activation into a graph Fourier basis and only retaining a fraction of the spectrum coefficients. In order to compute such an expansion more quickly, we consider applying the same method to each pair of layers separately (\subref{spectral_pair_layers}). This is less computationally expensive to compute but requires storing eigenvalues for each pair of layers, which is ultimately less space-efficient.}
\label{overview_figure}
\end{center}
\end{figure*}

\subsection{Activation Records}
\label{stats_section}
We now present details about the different types of activation records we used as strategies for reconstructing the original dataset. An overview of these techniques can be found in Figure \ref{overview_figure}.

\subsubsection{Top Layer Activation Statistics}
The simplest activation records we keep are the means and covariance matrices for each unit of the teacher's classification layer. This is also the layer used in conventional knowledge distillation \cite{hinton2015distilling}. We record these statistics according to Equation \ref{toplayer_stats_equation}, where $L$ refers to the values in the network right before the final softmax activation, $i$ refers to the $i$-th unit in that top layer, and $T$ refers to some temperature scaling parameter. In our experiments, we use temperature $T = 8$, just as in \cite{hinton2015distilling}.
\begin{equation}
\label{toplayer_stats_equation}
\mu_i = \text{Mean}({L_i}/{T}) \quad Ch_i = \text{Chol}(\text{Cov}({L_i}/{T}))
\end{equation}
To reconstruct the input, we first sample from these statistics and apply ReLU to it. We then replace the student's topmost non-linearity with ReLU and minimize MSE loss between these two activations by optimizing the input of the network, thus reconstructing an input that recreates the sampled activations.

\subsubsection{All Layers Activation Statistics}

Unfortunately the method above is underconstrained: there are many different inputs that can lead to the same top-layer activations, which means that our reconstructions aren't able to train the student model to very high accuracies. To better constrain the reconstructions, we store records for all layers instead of just the top-most. The reconstruction procedure is the same as the one above, except that for the hidden layers the statistics are as described in Equation \ref{alllayers_stats_equation}.
\begin{equation}
\label{alllayers_stats_equation}
\mu_i = \text{Mean}(L_i) \quad Ch_i = \text{Chol}(\text{Cov}(L_i))
\end{equation}
The optimization objective used was the sum of the MSE for each layer, normalized by the number of hidden units in the layer. This normalization is important in order to ensure that the relative importance of each layer is uniform.

However, simply reconstructing with statistics of all layers doesn't preserve inter-layer dynamics of chains of neurons that specialized together to perform some computation. In an attempt to preserve these, we freeze the dropout filters for each batch of reconstructed examples. This way, certain neurons will have their effect zeroed-out in each layer's activations, forcing other neurons to compensate. As can be seen in section \ref{results_section}, this addition was able to make the reconstructed set more visually similar to the original dataset. However, the student accuracies after data-free distillation were slightly worse than without the filters (see Table \ref{fc_accuracies_table}).

\subsubsection{Spectral Methods}
In order to better capture all of the interactions between layers of the network, we attempt to compress the entirety of the teacher network's activations rather than summarizing it statistically.

Many commonly used signal compression techniques are based around the idea of expanding a signal in some orthonormal basis, under the assumption that most of the information is stored in a small subset of those bases. If we represent the neural network's fully connected layers as a graph, and its activation as a graph signal, we can leverage the formulation of the Graph Fourier transform ($F_G$) presented in \cite{sandryhaila2013discrete} to compute a sparse basis for the activation of the network for a given class.

More formally, if we consider the neural network as the graph G(V, A), where V is a set of vertices corresponding to each neuron in a layer of the network and A is the adjacency matrix corresponding to the weighted connections of those neurons to each other, we can represent the network activation s as a real-valued graph signal \cite{sandryhaila2013discrete} on that network, which we can write as a vector $s= [s_0, s_1,...,s_{N-1}]^T \in \scriptsize{R}$ where each element $s_n$ is indexed by a vertex $v_n$ of the graph. We can then compress this graph signal by computing its graph Fourier basis and retaining only a small fraction of its largest spectrum coefficients. 

In the framework presented in \cite{sandryhaila2013discrete}, a graph Fourier basis simply corresponds to the Jordan basis of the graph adjacency matrix A: $A = V J V^{-1}$, where $F = V^{-1}$ is the graph Fourier transform matrix, with the frequency content $\hat{s}$ of s given by $\hat{s} = Fs$.

We can then compress the activation of the network by retaining only some fraction C of the spectrum coefficients $\hat{s}$ with the largest magnitude. Reconstructing the original signal is then done by simply inverting the initial Fourier transform matrix and multiplying by a zero-padded matrix of the spectrum coefficients: $\bar{s} = F_G^{-1}(\hat{s_0},..., \hat{s_C-1}, 0,...,0)^T$. Given an initial set of spectrum coefficients $\hat{s}$ and the corresponding graph Fourier transform matrix, we can compute the reconstruction loss for our network as the Euclidean distance between the reconstructed network activation $\bar{s}$ and $s_i$ the activation at a given iteration:
$$l = \sum_i (\bar{s} - s_i )^2 $$
The accuracy of the reconstructions is dependent on the number of retained spectrum coefficients, and so is less space efficient than the simpler statistical methods used above, but we find even just retaining $10\%$ of the spectrum coefficients yields a high degree of reconstruction accuracy. It is worth noting that it can be expensive to compute the eigendecomposition of large matrices, so we also consider reconstructing the teacher network activations based on spectra for the smaller graphs given by the connections of pairs of network layers instead of just using those of the entire graph.

\section{RESULTS}
\label{results_section}
Two datasets were chosen to examine different qualities of the proposed distillation method: MNIST \cite{lecun1998mnist}, which was used as a proof of concept that the proposed method works and to provide results that can be directly compared to Hinton et al. \cite{hinton2015distilling}, and CelebA \cite{liu2015faceattributes}, which was used to show that our method scales to large datasets and models.

For MNIST, we distilled a fully connected model and a convolutional model, to show versatility of the method. They were trained for 10 epochs using Adam with a learning rate of $0.001$. Any distillation procedures (including our method) were run for 30 epochs on the reconstructed datasets. Every reconstructed input was first initialized to per-pixel $\sim N(0.15, 0.1)$, and optimized using Adam.

For CelebA, we trained a larger convolutional model \cite{NIPS2012_4824}, to show scalability. We used learning rate $0.0001$ and input $\sim N(0.44, 0.29)$

We found that these input initializations, pixel means and variances of the training set, worked well for their respective datasets, and we posit that this kind of information could be provided as model metadata as well. Their computation can be done at training time with a running average of all pixels used for training, similar to what methods like Batch Normalization \cite{DBLP:journals/corr/IoffeS15} do.

\subsection{MNIST - Fully Connected Models}

For the experiments with fully connected models, we used the networks described by Hinton et al. \cite{hinton2015distilling}. A network comprising of two hidden layers of 1200 units (\textsc{Hinton-784-1200-1200-10}) was used as the teacher model, and was trained using dropout. Another network of two hidden layers of 800 units (\textsc{Hinton-784-800-800-10}) was used as the student. The total number of parameters was reduced by $50\%$. For each of them, the temperature parameter used was $8$, just like Hinton.

First, we trained both teacher and student models directly on MNIST. Then, we replicate the results from \cite{hinton2015distilling} by training the student model using knowledge distillation. The results can be found in Tables \ref{fc_accuracies_table} and \ref{fc_reconstructions_table}.

\begin{table*}[!ht]
\caption{Accuracies of the \textsc{Hinton} models and \textsc{MNIST} dataset for each procedure.} \label{fc_accuracies_table}
\begin{center}
\begin{tabular}{ l | l | c }
\textbf{Model} & \textbf{Activation Record} & \textbf{Accuracy on test set}\\
\hline
\textsc{Hinton-784-1200-1200-10} & Train on MNIST & $96.95\%$\\
\textsc{Hinton-784-800-800-10} & Train on MNIST & $95.70\%$\\
\hline
\textsc{Hinton-784-800-800-10} & Knowledge Distillation \cite{hinton2015distilling} & $95.74\%$\\
\hline
 & Top Layer Statistics & $68.75\%$\\
 & All Layers Statistics & $76.38\%$\\
\textsc{Hinton-784-800-800-10} & All Layers Statistics + Fixed Dropout Filters & $76.23\%$\\
 & All-Layers Spectral & $89.41\%$\\
 & Layer-Pairs Spectral & \textbf{$91.24\%$}\\
\end{tabular}
\end{center}
\end{table*}

{\def\arraystretch{1}\tabcolsep=3pt
\begin{table*}[!ht]
\caption{Per-class means and randomly sampled examples of datasets reconstructed using the different activation statistics from \textsc{Hinton-784-1200-1200-10}.} \label{fc_reconstructions_table}
\begin{center}
\begin{tabular}{ r c c }
\textbf{Activation Record} & \textbf{Means} & \textbf{Randomly sampled example}\\
\hline

MNIST & \includegraphics[width=0.37\textwidth]{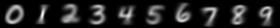} & \includegraphics[width=0.37\textwidth]{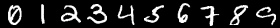}\\
\hline

Top Layer Statistics & \includegraphics[width=0.37\textwidth]{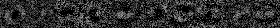} & \includegraphics[width=0.37\textwidth]{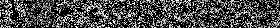}\\

All Layers Statistics & \includegraphics[width=0.37\textwidth]{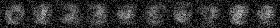} & \includegraphics[width=0.37\textwidth]{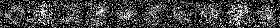}\\

All Layers + Dropout & \includegraphics[width=0.37\textwidth]{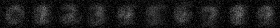} & \includegraphics[width=0.37\textwidth]{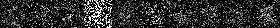}\\

Spectral All Layers & \includegraphics[width=0.37\textwidth]{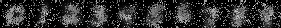} & \includegraphics[width=0.37\textwidth]{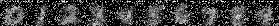}\\

Spectral Layer Pairs & \includegraphics[width=0.37\textwidth]{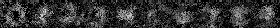} & \includegraphics[width=0.37\textwidth]{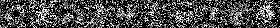}\\

\end{tabular}
\end{center}
\end{table*}
}


\subsection{MNIST - Convolutional Models}
For the experiments with convolutional models, we used \textsc{LeNet-5} \cite{lecun2015lenet} as the teacher model, and a modified version with half the number of convolutional filters per layer (\textsc{LeNet-5-half}) as the student model. The total number of parameters was reduced by $\sim 50\%$. The results can be found in Tables \ref{conv_accuracies_table} and \ref{conv_reconstructions_table}

\begin{table*}[!ht]
\caption{Accuracies of the \textsc{LeNet-5} model and \textsc{MNIST} dataset for each procedure.} \label{conv_accuracies_table}
\begin{center}
\begin{tabular}{ l | l | c }
\textbf{Model Name} & \textbf{Procedure} & \textbf{Accuracy on test set}\\
\hline
\textsc{LeNet-5} & Train on MNIST & $98.91\%$\\
\textsc{LeNet-5-half} & Train on MNIST & $98.65\%$\\
\hline
\textsc{LeNet-5-half} & Knowledge Distillation \cite{hinton2015distilling} & $98.91\%$\\
\hline
\textsc{LeNet-5-half} & Top Layer Statistics & $77.30\%$\\
 & All Layers Statistics & $85.61\%$\\
 & All-Layers Spectral  & $90.28\%$ \\
 & Layer-Pairs Spectral & $92.47\%$\\
\end{tabular}
\end{center}
\end{table*}

{\def\arraystretch{1}\tabcolsep=3pt
\begin{table*}[!ht]
\caption{Per-class means and randomly sampled examples of datasets reconstructed using the different activation statistics from \textsc{LeNet-5}.} \label{conv_reconstructions_table}
\begin{center}
\begin{tabular}{ r c c }
\textbf{Activation Record} & \textbf{Means} & \textbf{Randomly sampled example}\\
\hline

MNIST & \includegraphics[width=0.37\textwidth]{MEANS_og.png} & \includegraphics[width=0.37\textwidth]{RAND_og.png}\\
\hline

Top Layer Statistics & \includegraphics[width=0.37\textwidth]{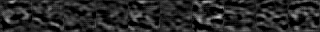} & \includegraphics[width=0.37\textwidth]{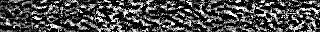}\\

All Layers Statistics & \includegraphics[width=0.37\textwidth]{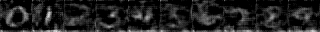} & \includegraphics[width=0.37\textwidth]{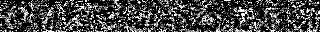}\\
Spectral All Layers & \includegraphics[width=0.37\textwidth]{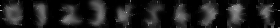} & \includegraphics[width=0.37\textwidth]{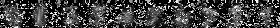} \\
Spectral Layer Pairs &  \includegraphics[width=0.37\textwidth]{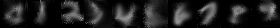}&  \includegraphics[width=0.37\textwidth]{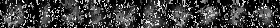}\\
\end{tabular}
\end{center}
\end{table*}
}


\subsection{CelebA - Convolutional Models}
In order to bring the experiments closer to the biometrics domain, and to show that our method generalizes to a larger task, we evaluate our approach on a model classifying the most balanced attribute in the large scale facial attributes dataset \textsc{CelebA} \cite{liu2015faceattributes} using the larger convolutional model \textsc{AlexNet} \cite{NIPS2012_4824}. As before, we use a student model \textsc{AlexNet-Half} with half the number of filters per convolutional layer.

As a note, we found that the All Layers optimization objective scales poorly with larger convolutional layers, as the covariance matrix grows at a much higher rate. Results for the other methods are shown in Table \ref{alex_conv_accuracies_table}.

\begin{table*}[!ht]
\caption{Accuracies of the \textsc{AlexNet} model and \textsc{CelebA} dataset for each procedure.} \label{alex_conv_accuracies_table}
\begin{center}
\begin{tabular}{ l | l | c }
\textbf{Model Name} & \textbf{Procedure} & \textbf{Accuracy on test set}\\
\hline
\textsc{AlexNet} & Train on CelebA & $80.82\%$\\ 
\textsc{AlexNet-half} & Train on CelebA & $81.59\%$\\ 
\hline
\textsc{AlexNet-half} & Knowledge Distillation \cite{hinton2015distilling} & $69.53\%$\\ 
\hline
\textsc{AlexNet-half} & Top Layer Statistics & $54.12\%$\\ 
 & All-Layers Spectral & $77.56\%$ \\
 & Layer-Pairs Spectral & $76.94\%$\\
\end{tabular}
\end{center}
\end{table*}


\section{DISCUSSION}
\label{discussion_section}

With the increasing popularity of deep learning methods requiring exorbitantly large numbers of parameters, it is useful to consider whether there may be better ways to distribute learned models. We have made the case that there may be metadata worth collecting at, or shortly after, training time that may facilitate the compression and distribution of these models.

However, different choices made about such metadata can have different tradeoffs with regards to the resulting compression accuracy and memory profile. The simple statistical methods presented in equations \ref{toplayer_stats_equation} and \ref{alllayers_stats_equation} are easy to compute and require little in the way of additional parameters, but suffer from limited compression accuracy even when reducing the overall memory of a profile by only $50\%$. Methods more similar to traditional image compression strategies (Figure \ref{spectral_all_layers}) require the retention of slightly more metadata in the form of vectors of spectral coefficients, yielding more accurate compression but being significantly more computationally expensive.

There are countless additional options that could be considered for such metadata, and we hope that this work can help spur some discussion into the development of standard deep neural network formats that may allow for their easier distribution. 

\section{CONCLUSION}

We have presented a method for data-free knowledge distillation. We have shown how many different strategies for activation recording can be used to reconstruct the original dataset, which can then be used to train the student network to varying levels of accuracy. We present tradeoffs related to the use of each of these strategies in section \ref{discussion_section}. We have shown that these activation records, if appended to the release of a pre-trained model, can facilitate its compression even in scenarios where the original training data is not available.

\bibliography{bib}{}
\bibliographystyle{plain}

\end{document}